\documentclass[letterpaper, 10 pt, conference]{ieeeconf}  

\IEEEoverridecommandlockouts                              

\usepackage{graphics} 
\usepackage{adjustbox}
\usepackage{epsfig} 
\usepackage{amsmath} 
\usepackage{amssymb}  
\usepackage{xcolor}
\usepackage{cite}
\usepackage{booktabs}
\usepackage{caption}
\usepackage{diagbox}
\usepackage{multirow}
\usepackage{booktabs}
\usepackage{makecell}
\usepackage{array} 
\usepackage{xcolor,graphicx}
\usepackage{array} 
\usepackage{url}
\newcolumntype{P}[1]{>{\centering\arraybackslash}p{#1}}

\newcommand{\eg}{\textit{e.g.}}

\newcommand{\faraz}[1]{{\ifnum\Comments=1\textcolor{purple}{[faraz: #1]}\fi}}

\title{\LARGE \bf
LEACL: LLM-Enhanced Automatic Curriculum Learning for Reinforcement Learning in Long-Horizon Manipulation Tasks
}

\author{
Faraz Heravi$^{1*}$, James Ouyang$^{1*}$, Zifan Xu$^{1}$, Arjun Kumar$^{1}$, Yoonchang Sung$^{2}$, Peter Stone$^{1,3}$
\thanks{$^{*}$ detonates equal contribution, $^{1}$The University of Texas at Austin,
        {\tt\small \{farazh, jouyang, zfxu, arjunk30\}@utexas.edu, pstone@cs.utexas.edu}. $^{2}$Nanyang Technological University, {\tt\small yoonchang.sung@ntu.edu.sg}. $^{3}$Sony AI. This work has taken place in the Learning Agents Research
Group (LARG) at UT Austin.  LARG research is supported in part by NSF (FAIN-2019844, NRT-2125858), ONR (N00014-18-2243), ARO (E2061621), Bosch, Lockheed Martin, and UT Austin's Good Systems grand challenge.
Peter Stone serves as the Executive Director of Sony AI America and receives financial compensation for this work.  The terms of this arrangement have been reviewed and approved by the University of Texas at Austin in accordance with its policy on objectivity in research.}
}

\begin{document}

\maketitle
\thispagestyle{empty}
\pagestyle{empty}

\begin{abstract}
Long-horizon manipulation tasks pose significant challenges for reinforcement learning due to sparse reward signals and long horizons. Automatic curriculum learning (ACL) has been proposed to tackle these challenges by progressively training agents on a sequence of tasks, from easier to more difficult. However, the success of ACL depends heavily on task-dependent specifications—such as well-defined task parameter spaces and difficulty measures—which are often manually crafted and difficult to generalize across diverse tasks. Recent advances in large language models (LLMs) offer a promising alternative by enabling the decomposition of complex tasks into meaningful subtasks using the LLMs' web-scale common-sense knowledge. This decomposition can provide a natural curriculum structure for efficient learning of long-horizon tasks. However, existing LLM-based methods typically rely on hand-designed dense reward functions to learn each subtask, which can introduce bias and still requires significant human supervision. 

In this work, we propose LLM-enhanced automatic curriculum learning (LEACL), a framework that integrates LLMs and ACL to address these limitations. Specifically, LLMs are used to both decompose tasks into subtasks and to generate task-dependent specifications for each subtask. These specifications are then used by ACL algorithms to guide learning using only sparse reward signals, eliminating the need for dense reward design. We evaluate LEACL on five long-horizon manipulation tasks from the LIBERO benchmark. LEACL achieves better asymptotic performance in terms of the success rates compared to human-designed dense rewards.
\end{abstract}


\section{Introduction}

\begin{figure*}[htb!]
    \centering
    \includegraphics[width=1.0\linewidth]{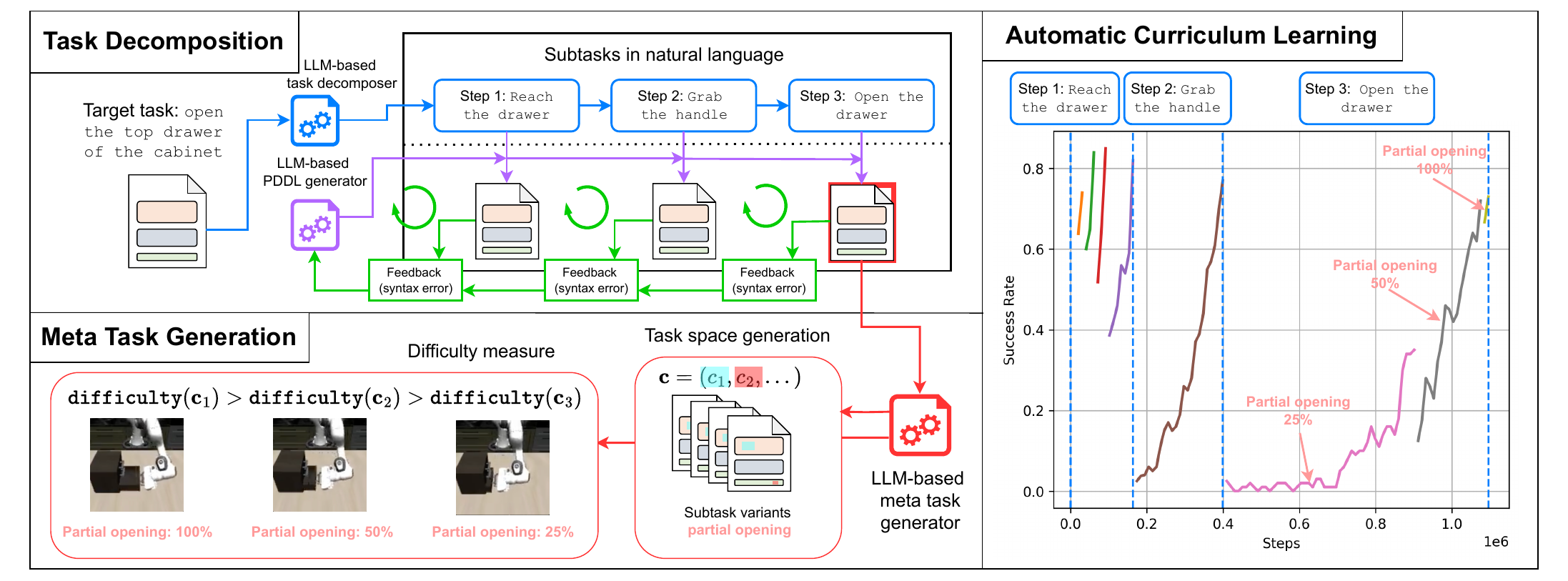}
    \caption{LEACL stages: (1) LLM-based task decomposition; (2) LLM-based meta-task generation; (3) Automatic CL.}
    \label{fig:lecl_overview}
    \vspace{-10pt}
\end{figure*}

Reinforcement learning (RL) has begun to show promise in solving manipulation tasks that require multiple interactions with objects to achieve specific goals 
\cite{plappert2018multi, andrychowicz2020learning, tang2024deep}. For example, \emph{turning on a stove} may require the agent to learn behaviors composed of reaching, grasping, rotating, and applying force—behaviors that typically emerge only with carefully designed dense rewards to guide learning. However, manipulation involving complex dynamics (\eg, maintaining precise contact or alignment) in long-horizon tasks remains particularly challenging. Designing effective reward functions for such tasks and integrating multiple complex rewards to enable the learning of extended action sequences are difficult and often require substantial manual effort. While \emph{sparse task completion rewards} eliminate the need for manual reward specification, they often lead to high sample complexity due to the inherent challenges of exploration in high-dimensional, continuous state spaces over long horizons.

To improve sample efficiency in these sparse-reward settings, automatic curriculum learning (ACL) has emerged as a powerful approach \cite{portelas2020automatic, narvekar2020curriculum,romac2021teachmyagent}. Instead of training solely on the target task, ACL dynamically adjusts the training task distribution by gradually transitioning from simpler tasks, where learning progress is easier, to more difficult ones. Methods like APT-Gen \cite{fang2020adaptive} have shown the potential of ACL in robotic manipulation. However, these methods often rely on hand-crafted task-dependent specifications, such as a \emph{task parameter space} and  \emph{difficulty measures}. For example, in a task requiring a robot to grasp an object, a useful task parameter space might include variables that specify the initial pose of the gripper and the initial placement of the object. Designing such parameter spaces is particularly difficult when the space of relevant object configurations and interactions is high-dimensional. Other task-dependent specifications, such as difficulty measures, can substantially impact performance but are likewise challenging to define manually~\cite{romac2021teachmyagent}.

Another direction exploits task-specific structures for curriculum design, often leveraging task and motion planning (TAMP) frameworks to decompose complex tasks into simpler subtasks \cite{garrett2021integrated,cheng2023league}. For example, \emph{placing a plate in a closed microwave} can be decomposed into a sequence of four subtasks: (1) \textit{open the microwave}, (2) \textit{pick up the plate}, (3) \textit{place it inside}, and (4) \textit{close the door}.
While these approaches effectively decompose tasks by identifying subgoals, the logical expressions for task-specific structures are often hand-engineered, and dense reward functions are still required for each subtask—especially those involving dexterous interactions, such as maintaining contact or achieving precise alignment. 

Recent advances in large language models (LLMs) offer a promising alternative that sidesteps the challenges of introducing TAMP-driven task-specific structures. By leveraging their web-scale common-sense knowledge, LLMs can generate the logical expressions required for task decomposition~\cite{liu2023llm+,li2024league++}, as well as the corresponding reward functions, directly from natural language descriptions~\cite{yu2023language,li2024league++,ma2023eureka,ma2406dreureka}. However, many existing LLM-based reward generation methods still rely on predefined reward primitives or manually designed reward templates, limiting their flexibility across diverse tasks. Furthermore, dense reward functions may misalign with the true sparse objectives, potentially leading to task failure.

To develop a general-purpose and effective approach for learning long-horizon manipulation tasks without task-specific human input, this work introduces LLM-enhanced automatic curriculum learning (LEACL)—a framework that integrates LLM-based curriculum generation, simplified through TAMP structures, with ACL methods while eliminating the need for manually designed dense reward functions. 

LEACL operates in three key stages: (1) \textbf{LLM-based task decomposition}, which breaks down long-horizon tasks into structured subgoals using LLMs, similar to LEAGUE++~\cite{li2024league++}; (2) \textbf{LLM-based meta-task generation}, which generates task-dependent specifications for each subtask, guiding automatic curriculum generation; (3) \textbf{automatic curriculum learning}, which applies well-established ACL algorithms to generate curricula using the task-dependent specifications provided by the LLM.

The paper makes the following main contributions.
\begin{itemize}
    \item We propose a novel LLM-enhanced automatic curriculum learning framework, LEACL, which enables RL agents to solve challenging long-horizon manipulation tasks using only sparse task completion rewards, without any hand-crafted effort.

    \item Empirical validation demonstrates that LEACL achieves superior sample efficiency and higher success rates after convergence compared to baselines without ACL, while delivering competitive performance relative to a baseline that relies on task-specific human engineering for task decomposition and meta-task generation. 
\end{itemize}
\section{Related Work}
The main strength of LEACL stems from two key principles: (1) leveraging LLMs to eliminate the need for human input, and (2) incorporating CL on decomposed subtasks to address sparse rewards. We briefly review related work in these two areas. 

\subsection{LLMs for planning and decision making}
LLMs have significantly advanced robotic planning and decision-making, as demonstrated by a growing body of research \cite{ahn2022can,rana2023sayplan,singh2023progprompt,huang2022language}. These studies showcase LLMs' capabilities in interpreting complex natural language commands and generating coherent action plans, thereby pushing robotic autonomy forward with minimal human supervision. However, using LLMs to directly produce low-level motion plans remains a substantial challenge due to the gap between symbolic reasoning and physical execution. Liang et al. \cite{liang2023code} explore the ``code-as-policy" paradigm, where LLMs directly generate scripted control policies. While this approach is interpretable and allows for human-readable plans, it often struggles with execution-time robustness and generalization to dynamic or uncertain environments. 

Another line of work combines Reinforcement Learning (RL) with LLMs. For example, LEAGUE++ \cite{li2024league++} uses LLMs for automatic task decomposition, symbolic operator creation, and dense reward generation to guide the learning of each symbolic action. Similarly, Text2Reward \cite{xie2024textreward} employs LLMs to synthesize dense reward functions directly from natural language task descriptions. However, LEACL instead uses LLMs to generate structured task specifications and parameterized task spaces for ACL, avoiding the need to design or generate dense reward functions. More recent progress has explored increasingly sophisticated approaches to programmatic reward generation. For instance, ARCHIE \cite{turcato2025towards} focuses on generating rewards for learning complex, contact-rich single skills, while CurricuLLM \cite{ryu2025curricullm} adopts a hierarchical strategy where an LLM-generated curriculum guides subsequent reward function creation for each subtask. Rather than generating dense reward functions, LEACL generates structured task specifications and parameterized subtasks that enable ACL algorithms to learn directly from sparse rewards. While there is no direct one-to-one comparison between ACL and LLM-based reward generation, we provide results from the LEAGUE baseline, which uses human-expert reward functions as a surrogate. Pushing into multimodality, IKER \cite{patel2025real} leverages Vision Language Models (VLMs) to derive rewards from visual keypoints, though it focuses on chaining separately learned skills rather than learning a single policy. Its primary challenge is robust skill chaining, whereas LEACL focuses on learning a unified RL policy. While these methods highlight advanced techniques for reward design, they underscore a common dependency on crafting dense reward signals. In contrast, our method also leverages LLMs’ high-level planning abilities to decompose the tasks into sub-tasks and uses RL for learning low-level motion plans. However, unlike these approaches, LEACL sidesteps reward engineering entirely, using LLM-derived priors to structure an efficient task space that enables learning directly from sparse rewards with ACL. This allows for more scalable and adaptable skill acquisition across diverse robotic tasks.

\subsection{Curriculum learning for RL}
Curriculum learning (CL) has emerged as a widely adopted training paradigm to improve sample efficiency in RL, particularly for hard-exploration problems. CL typically involves training an RL agent on a sequence of source tasks—progressing from easier to harder—until the agent successfully learns a desired target task. While early CL approaches relied on manually designed task sequences \cite{narvekar2020curriculum}, recent advances in ACL have introduced methods that adaptively adjust the task distribution during training \cite{portelas2020automatic}. However, these ACL algorithms are often sensitive to task-specific configurations, including the parameterization of tasks and difficulty measures.

On the other hand, recent work has explored the use of LLMs for curriculum design. For instance, Wang et al. \cite{wang2023voyager} employ LLMs to propose achievable yet challenging goals for agents in the Minecraft games. 
Eurekaverse \cite{liang2024eurekaverse} utilizes LLMs to generate emergent parkour environments for learning quadrupedal locomotion.
While these approaches are effective, they often rely heavily on handcrafted prompts that explicitly instruct the LLM on which tasks to select under specific conditions. In contrast, LEACL proposes using LLMs as meta-task generators, which generate task-dependent specifications, instead of directly generating curricula. This strategy not only simplifies prompt design and reduces the number of LLM queries required during training, but also enables the use of well-established ACL methods.
\section{Problem Description}
\label{sec:prob}
\begin{figure}
    \centering
    \includegraphics[width=0.9\linewidth]{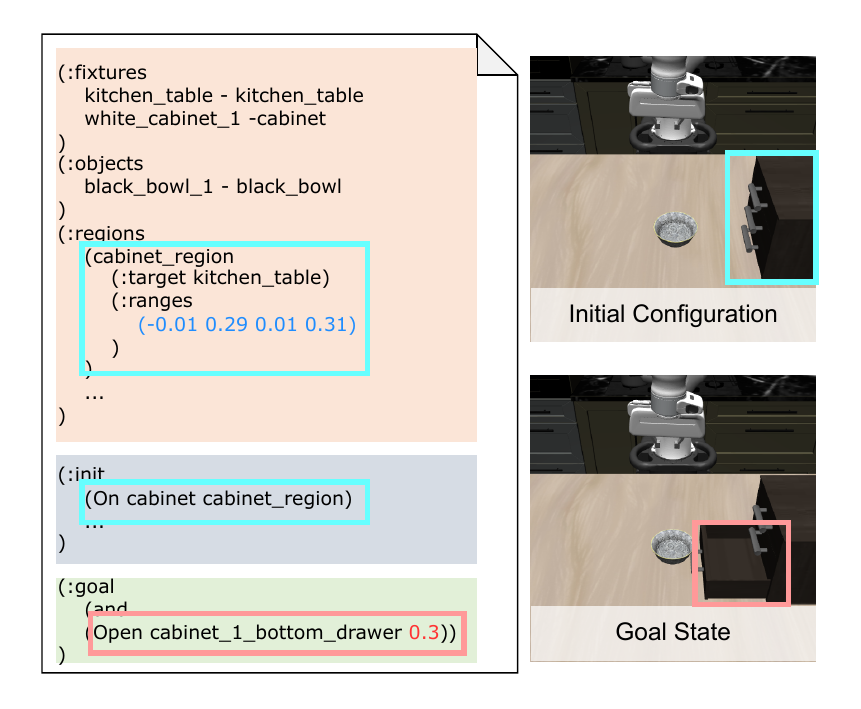}
    \caption{An example PDDL specification of the task \texttt{open the bottom drawer of the cabinet}. The \textcolor[HTML]{67ffff}{blue} and \textcolor[HTML]{ff9999}{red} blocks highlight how initial state and goal state can be controlled respectively.}
    \label{fig:PDDL_task}
    \vspace{-10pt}
\end{figure}

In this work, we consider an environment composed of $\mathcal{M}, \mathcal{F},$ and $\mathcal{R}$, denoting a set of movable objects (\eg, bowls and cups), a set of fixtures (\eg, kitchen tables and walls), and a set of regions (\eg, a cabinet region where a movable object can be placed), respectively. The robot is capable of grasping a movable object and placing it in a region on a fixture.

We assume full observability of the environment, allowing manipulation tasks to be formulated as Markov decision process (MDP) problems. In particular, we consider a finite-horizon, goal-augmented MDP: $M =
\langle \mathcal{S}, \mathcal{A}, T, \mu, R, \mathcal{S}_G \rangle$. Here, $\mathcal{S}$ represents the state space of the robot (\eg, configurations) and environment (\eg, poses of $\mathcal{M}, \mathcal{F},$ and $\mathcal{R}$), while $\mathcal{A}$ denotes the action space of the robot (\eg, end-effector pose). The transition function is defined as $T \colon \mathcal{S} \times \mathcal{A} \rightarrow \mathcal{S}$, $\mu$ represents the initial
state distribution, the reward function is given by $R \colon \mathcal{S} \rightarrow \mathbb{R}$, and the goal set $\mathcal{S}_G \subset \mathcal{S}$ denotes the subset of states where the task is considered complete.

The types of tasks considered in this work require the robot to reposition movable objects from one region to another, involving contact-rich interactions with fixtures such as opening a drawer, closing a microwave, and turning on a stove.

In this work, we consider a sparse-reward setting, where the reward is given only upon task completion, with $R=1$ if the state is in $\mathcal{S}_G$, and $R=0$ otherwise. The robot terminates its learning episode upon task completion, making the maximum obtainable reward $1$. 

The objective of this work is to learn a policy $\pi$ that maximizes the expected return, given natural language describing the goal of a task as input: $\max_\pi J(\pi) = \mathbb{E}_{\mathcal{A}\ni a_t \sim \pi, \mu} [
\sum_{t=1}^H R(s_t\in\mathcal{S})]$, where $H$ is the finite horizon. 

\section{LEACL}
\label{sec:leacl}
\begin{figure*}[htb!]
    \centering
    \includegraphics[width=0.95\linewidth]{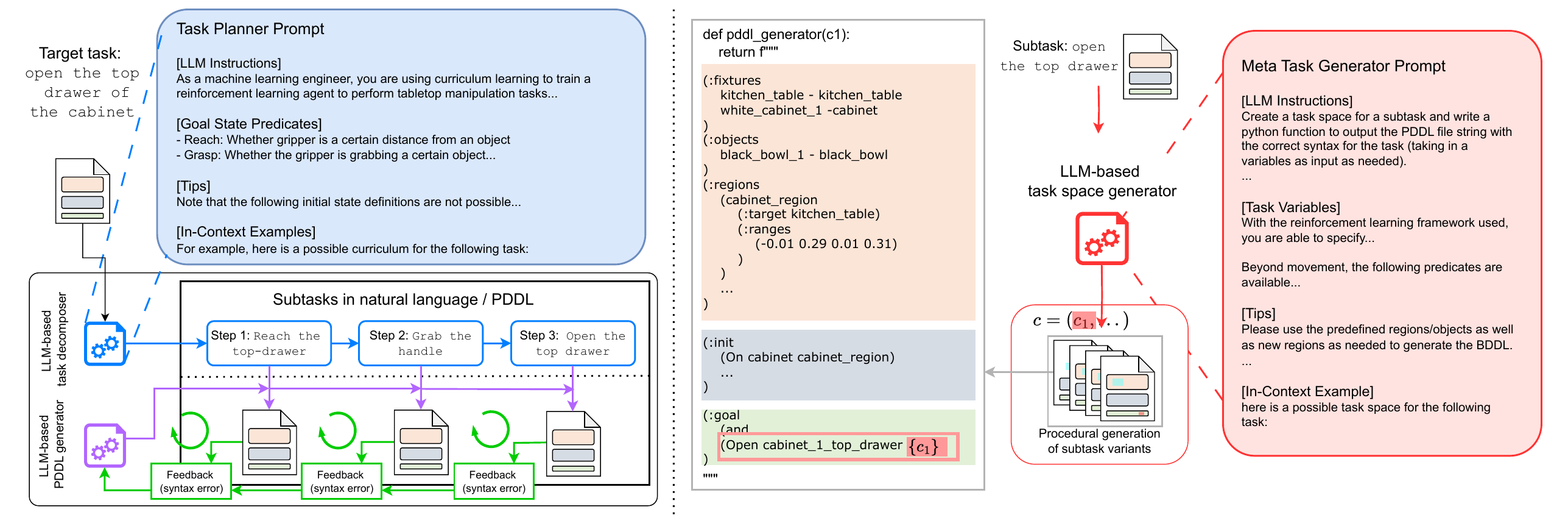}
    \caption{The prompts used for LLM-based task decomposition (left) and meta-task generation (right).}
    \label{fig:task_planner}
    \vspace{-10pt}
\end{figure*}
Our LEACL framework comprises three key stages executed sequentially (Fig.~\ref{fig:lecl_overview}):
\begin{itemize}
    \item LLM-based task decomposition, which breaks down a long-horizon manipulation task into a sequence of semantically meaningful and manageable subtasks.
    \item LLM-based meta-task generation, which generates task-dependent specifications to facilitate ACL.
    \item Plug-and-play ACL, which ultimately solves a given sparse-reward task using the information generated above.
\end{itemize}
Below, we describe each stage in detail.

\subsection{LLM-based task decomposition}
Long-horizon manipulation tasks often suffer from sparse rewards, making them difficult to solve using standard RL methods. To mitigate this complexity, we draw inspiration from classical planning approaches that decompose complex tasks into semantically meaningful subtasks, each providing intermediate learning signals~\cite{gehring2022reinforcement}—a strategy adopted by LEACL in its first stage. 

This decomposition leverages the rich reasoning capabilities of LLMs to produce a sequence of subtasks, each encoded as a logical specification written in the planning domain definition language (PDDL~\cite{aeronautiques1998pddl}) specification (Fig. \ref{fig:PDDL_task}), inspired by TAMP research. A grammar is required to define PDDL specifications as building blocks, referred to as predicates (denoted as goal state predicates in Fig.~\ref{fig:task_planner} (left)). We assume that the complete predicate vocabulary available in the target domain (e.g. LIBERO+) is provided to the LLM, from which it selects the predicates needed for the given task.

As shown in Fig.~\ref{fig:task_planner} (left), the task decomposition process follows a bi-level prompting strategy. At the high level, given a goal description in natural language—referred to as the target task—an LLM generates a sequence of subtasks, each representing a single-action task described in natural language. Each natural language subtask description is then passed to a second LLM, which converts it into a valid PDDL specification using structured prompts. This process incorporates reflection mechanisms~\cite{pan2023automatically,kambhampati2024position}, such as syntax validation and iterative refinement, to improve syntactic validity and executability of the generated PDDL specifications.

Let $K$ denote the number of subtasks, generated and determined dynamically by the LLM. We denote the corresponding sequence of $K$ PDDL specifications as $(\tau_i)_{i=1}^K$, where each specification is represented as: $\tau_i= \langle \mathcal{M}, \mathcal{F}, \mathcal{R}, \mathcal{P}_i, \mathcal{I}_i, \mathcal{G}_i \rangle \in \mathcal{T}$, where:
\begin{itemize}
\item $\mathcal{M}, \mathcal{F},$ and $\mathcal{R}$ denote components of the environment as defined in Section~\ref{sec:prob}.
\item $\mathcal{P}_i$ denotes a set of predicates, where each predicate is a Boolean function evaluated on a tuple of movable objects, fixtures, regions, and/or typed variables to determine whether it is true or false. An assignment of values to a predicate is called a literal. For instance, \texttt{Open}(drawer$\in\mathcal{F}$, $0.3$) is true if the drawer is opened by more than $0.3m$. The set of predicates, $\cup_{i=1}^K \mathcal{P}_i$, defines a grammar.
\item $\mathcal{I}_i$ and $\mathcal{G}_i$ denote a set of initial literals and the conjunctive set of goal literals, respectively.
\end{itemize}

Note that we do not introduce an index $i$ for movable objects, fixtures, and regions, meaning that all subtasks share them consistently. This constraint is explicitly encoded in the LLM prompt. Additionally, note that the action (or operator) term is omitted, as each specification corresponds to a degenerate, single-action task, making it unnecessary.

Each specification $\tau_i$ is paired with a manipulation subtask $M_i$, modelded as an MDP, where the initial state distribution $\mu$ and the goal set $S_G$ are defined such that the start state satisfies $\mathcal{I}_i$ and the goal state satisfies $\mathcal{G}_i$, respectively. As a result of this stage, we obtain a sequence of subtask–PDDL specification pairs, denoted as $(\langle M_i, \tau_i \rangle)_{i=1}^K$.


\subsection{LLM-based meta-task generation}
\label{sec:meta_gen}
While task decomposition simplifies exploration by introducing intermediate goals, learning policies from sparse rewards for each subtask still remains challenging due to the complexity of contact-rich subtasks. Prior work~\cite{li2024league++} has attempted to prompt LLMs to generate dense reward functions, but such heuristics often lack generality across diverse manipulation tasks. 
Instead, LEACL uses LLMs to generate task-dependent specifications that facilitate the use of ACL methods, including task parameter space and difficulty measure. These task-dependent specifications are generated as follows.


\paragraph{Task parameter space} Given a decomposed subtask, the LLM-based meta-task generator first defines the corresponding parametric task space based on the PDDL specification $\tau_i$. This involves the LLM identifying a $N_i$-dimensional parameter vector $\mathbf{c}_i = (c_i^j)_{j=1}^{N_i}$, which form a control space $\mathcal{C}_i$ that can be derived from the state space $\mathcal{S}$. Each vector $\mathbf{c}_i \in \mathcal{C}_i$ parameterizes an instance of the specification $\tau_i$. For example, in the subtask \textit{open the top drawer of the cabinet}, the control variables may include the cabinet’s initial position and the target drawer opening distance. The LLM then effectively defines a generation function that maps the control space $\mathcal{C}_i$ to a subset of valid PDDL specifications $\mathcal{T}_i \subset \mathcal{T}$. This mapping is realized by the LLM generating a Python function template (as exemplified in Fig.~\ref{fig:task_planner}, right panel).

\paragraph{Difficulty measure} 
Difficulty measure is common meta-task information for curriculum generation as progressively training from easy to hard tasks usually leads to more efficient learning of the hard tasks. To incorporate the difficulty measure in the meta-task generation, a LLM is prompted to generate a sequence of $D$ parameter vectors $(\mathbf{c}_i^1, ..., \mathbf{c}_i^D)$ that is explicitly ordered by difficulty. This ordering implies that for any $p,q \in \{1,…,D\}$ where $p > q$, the task $\mathbf{c}_i^p$ is more difficult than $\mathbf{c}_i^q$ based on the difficulty measure of LLMs. 

\subsection{Automatic curriculum learning (ACL)}
ACL aims to improve RL efficiency by adaptively selecting training tasks based on the agent’s current capabilities. Specifically, for each subtask $M_i$ (obtained via LLM-based task decomposition) and its associated LLM-generated task parameter space $\mathcal{T}_i$. ACL then maps the agent's training history—such as its past episodic rewards (ALP-GMM~\cite{portelas2020teacher}) or temporal difference errors (PLR~\cite{jiang2021prioritized})—to a distribution over tasks within $\mathcal{T}_i$. The optimization objective of an ACL method can be summarized as maximizing the performance (\eg, success rate) of a post-trained agent on the original subtask $M_i$.

Depending on the ACL algorithm, task-dependent specifications may or may not be required. Table \ref{tab:priors} summarizes the dependence on task-dependent specifications of recent ACL algorithms.
\begin{table}[h]
    \centering
    \renewcommand{\arraystretch}{1.2}
    \begin{tabular}{lccc}
        \toprule
        \textbf{ALGORITHM} & Task parameter space & Difficulty measure \\
        \midrule
        ADR \cite{mehta2020active} & OPT. & REQ.  \\
        ALP-GMM \cite{portelas2020teacher} & OPT. & OPT. \\
        COVAR-GMM \cite{moulin2014self} & OPT. & OPT. \\
        GOAL-GAN \cite{florensa2018automatic} & OPT. & OPT.  \\
        RIAC \cite{baranes2009r} & OPT. & - \\
        SPDL \cite{klink2020self} & OPT. & REQ. \\
        SETTER-SOLVER \cite{racaniere2019automated} & OPT. &  OPT. \\
        \bottomrule
    \end{tabular}
    \caption{Dependence on task-dependent specifications priors. OPT. and REQ. are short for optional and required, respectively.}
    \label{tab:priors}
\end{table}

The final component of LEACL connects with standard ACL algorithms to train agents on the generated subtask parameter spaces. LEACL adopts a plug-and-play design compatible with many ACL approaches through standardized interfaces. In particular, we build upon the TeachMyAgent framework \cite{romac2021teachmyagent}, which supports a variety of ACL methods. 

Finally, a sequence of $K$ subtasks, successively solved by the selected ACL algorithm, is aggregated to form a solution to the given manipulation target task.

\begin{table*}[htb!]
\centering
\resizebox{\linewidth}{!}{%
\begin{tabular}{l|P{2.5cm}|P{2cm}|P{2.8cm}|P{2.8cm}|P{2.7cm}} 
\toprule
\multirow{2}{*}{\diagbox{Methods}{Tasks}} & \centering\textit{Open the bottom drawer of the cabinet} & \centering\textit{Put the white bowl on the plate} & \centering\textit{Pick up the ketchup and put it in the basket} & \centering\textit{Turn on the stove and put the moka pot on it} & \textit{Put the mug in the microwave and close it}  \\
\midrule
 Sparse reward               & 0.0 $\pm$ 0.0 & 0.0 $\pm$ 0.0 & 0.0 $\pm$ 0.0 & 0.0 $\pm$ 0.0 & 0.0 $\pm$ 0.0   \\
LEACL w/o ACL     & 0.0 $\pm$ 0.0 & 11.8 $\pm$ 15.4 & 0.0 $\pm$ 0.0 & 8.2 $\pm$ 15.5 & 0.0 $\pm$ 0.0 \\
LEAGUE \cite{cheng2023league}                      & 99.4 $\pm$ 1.0 & 71.0 $\pm$ 37.4 & 29.8 $\pm$ 50.6 & 21.1 $\pm$ 17.4 & 0.0 $\pm$ 0.0 \\
Human curriculum  & 99.8 $\pm$ 0.2 & \textbf{96.0 $\pm$ 2.5} & 86.3 $\pm$ 7.2 & \textbf{79.7 $\pm$ 11.1} & \textbf{89.0 $\pm$ 7.5} \\
LEACL (ours)                        & \textbf{99.8 $\pm$ 0.1} & 90.7 $\pm$ 4.1 & \textbf{89.4 $\pm$ 1.8}   & 60.6 $\pm$ 6.5 & 75.9 $\pm$ 3.4    \\
\bottomrule
\end{tabular}}
\caption{The success rates (\%) of the evaluated approaches on five manipulation tasks. Results are reported as mean $\pm$ 95\% confidence interval across five independent random seeds. Each seed is evaluated over 1000 episodes using the final model. Bold numbers indicate the best-performing algorithms.}
\label{tab:results}
\end{table*}

\section{Experiments}
Our experimental contributions are twofold. First, we utilize the existing LIBERO benchmark~\cite{liu2024libero}, developed for lifelong learning in robot manipulation, which exhibits limitations when applied to CL settings. We extend its functionality to support CL algorithms, resulting in LIBERO+, which is publicly available.\footnote{\url{https://github.com/fheravi/LIBERO-plus}.} Second, we compare the performance of LEACL with baselines that either omit some components of LEACL or use human-designed components, across five manipulation tasks generated in LIBERO+, demonstrating higher success rates.

\subsection{LIBERO+}
\label{sec:LIBERO+}
\begin{figure*}[htb!]
    \centering
    \includegraphics[width=0.95\linewidth]{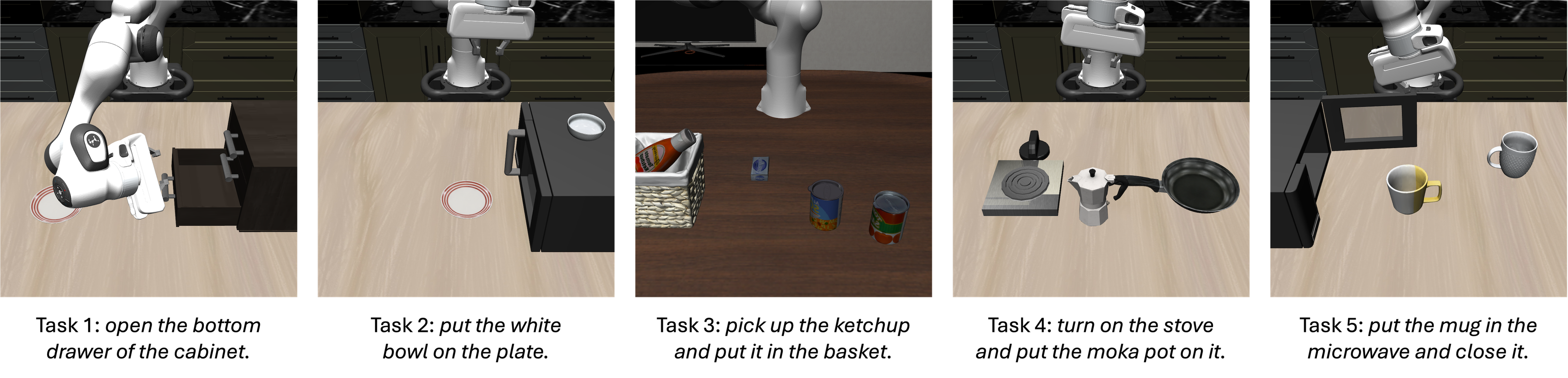}
    \caption{Front camera views of the five evaluation tasks.}
    \label{fig:enter-label}
    \vspace{-10pt}
\end{figure*}
While our approach is broadly applicable to general PDDL-based procedural task generation pipelines, in this paper, we focus on LIBERO, a benchmark originally introduced to evaluate knowledge transfer in lifelong robot learning. LIBERO provides a rich set of movable objects, fixtures, and predicate definitions, along with corresponding MuJoCo~\cite{todorov2012mujoco} simulation assets. 

To support finer-grained task generation, we extend the original LIBERO with an additional set of parameterized predicates, referring to this enhanced version as LIBERO+~\cite{bar2025thesis}. This extension is crucial for enabling task decomposition and effective ACL. Table~\ref{tab:components} shows the different components of LIBERO+ used by LEACL and baseline algorithms. 

Task diversity is crucial for CL. In the original PDDL-based procedural generation pipeline, task definitions are constrained by the combinations of valid PDDL syntax—particularly the predicates that specify initial and goal states. LIBERO defines a set of \emph{unary} predicates to describe object states and \emph{binary} predicates to express spatial relationships.

LIBERO+ extends LIBERO by replacing predicates that rely on certain parameters with their \emph{parameterized} counterparts. For example, the predicate \texttt{open(object)} can be replaced with \texttt{open(object, ?o)}, which includes an additional continuous-valued parameter \texttt{o} specifying the degree to which the object is opened. Furthermore, LIBERO+ introduces several additional predicates to support more fine-grained task generation. In summary, LIBERO+ enables more diverse tasks and PDDL specifications than LIBERO by introducing new predicates and supporting continuous-valued parameters.

\subsection{Evaluation tasks}
We present the five manipulation tasks used for evaluation below. These target tasks are referred to by their natural language descriptions:
\begin{itemize}
    \item Task 1: \textit{open the bottom drawer of the cabinet}. 
    \item Task 2: \textit{put the white bowl on the plate}.
    \item Task 3: \textit{pick up the ketchup and put it in the basket}.
    \item Task 4: \textit{turn on the stove and put the moka pot on it}.
    \item Task 5: \textit{put the mug in the microwave and close it}. 
\end{itemize}
These tasks vary in complexity and involve diverse object interactions and goal configurations. For evaluation purposes, each task can be decomposed by a human expert into a sequence of subtasks. The number of subtasks for Tasks 1 through 5 are 2, 5, 6, 4, and 5, respectively.

The five tasks share a common action space but differ in their state spaces, which are composed of low-dimensional representations of task-relevant object states, the end-effector’s 3D position and orientation, the robot’s joint positions and velocities, and the goal state. For example, in the task \textit{turn on the stove and put the moka pot on it}, the task-relevant object states include the 3D position and orientation of both the moka pot and the stove, represented relative to the robot’s end-effector.

For each task, the evaluated approach trains five independent policies using a predetermined training budget.\footnote{The training budget is chosen to allow all baseline approaches to reach their success rate plateaus, with a minimum of 500k environment steps.} The average success rates are reported across the five trained policies, estimated from 1,000 rollout trajectories.

\subsection{Baselines}
LEACL is evaluated against the following baselines. Table~\ref{tab:components} summarizes the differences among these algorithms. 
\paragraph{Base RL algorithm}
We utilize Proximal Policy Optimization (PPO) \cite{schulman2017proximal} as the base RL algorithms in the experiments. Hyperparameters are adapted mainly from Stable-Baselines3 \cite{stable-baselines3}. However, we adjusted a few key parameters based on informal empirical analysis on short-horizon manipulation tasks, including closing the microwave door and turning on the stove, using only sparse reward and no task decomposition, in order to improve training stability, characterized by consistently converging average reward and task completion rate within a limited number of time steps. Specifically, we modified the learning rate to 0.0002 and set the number of environment steps per policy update to 512. We used a multi-layer perceptron (MLP) architecture with two hidden layers of 128 units each for both the actor and critic networks. These hyperparameters were held constant across all tasks to ensure consistent and fair comparisons.

\paragraph{Sparse reward} This baseline trains a policy using only the final task completion reward, without any intermediate or shaped rewards, and without using ACL approaches.

\paragraph{LEACL w/o ACL} This baseline trains a policy on a sequence of subtasks that are manually defined by human experts. For each subtask, only a sparse task completion reward is provided with no shaping rewards.

\paragraph{LEAGUE \cite{cheng2023league}} As a baseline, we adapt the LEAGUE framework \cite{cheng2023league}, where the agent learns a sequence of subtasks guided by expert-designed dense reward functions. We manually crafted effective shaping rewards for each subtask in our evaluation suite to ensure reliable skill acquisition for each subtasks. We consider this expertly-tuned baseline a strong surrogate for automated reward generation approaches, as the handcrafted functions likely represent an upper bound on the performance achievable by a generated reward policy.

\paragraph{Human curriculum} This baseline mirrors the three-stage pipeline of LEACL, but the task decomposition and task-relevant priors are fully designed by human experts\footnote{Human experts are the authors of the paper who have full knowledge of the tasks and are experienced at teleoperating the manipulation arm and at designing curricula for RL training.} instead of LLMs, and is thus expected to provide an upper-bound on performance among the evaluated algorithms.

\paragraph{LEACL} Our proposed method, described in Section~\ref{sec:leacl}, uses ChatGPT 4o-mini \cite{openai2024gpt4omini} as the LLM-based task decomposer, PDDL generator, and meta-task generator. Although LEACL can be integrated with various ACL algorithms, we use active domain randomization \cite{mehta2020active} as the downstream ACL algorithm. While simple, we found it effective for solving the sub-tasks involved in this work if proper task-relevant priors are provided.

\begin{table}[h]
    \centering
    \renewcommand{\arraystretch}{1.2}
    \setlength{\tabcolsep}{2pt}
    \begin{tabular}{lccc}
        \toprule
        \textbf{Baselines} & \multicolumn{1}{p{1.7cm}}{\centering Task\\Decomposition} & \multicolumn{1}{p{1.7cm}}{\centering Parameterized\\Predicates} & \multicolumn{1}{p{2cm}}{\centering Specifications\\in RL}  \\
        \midrule
        Sparse reward & $\times$ & $\times$ & Sparse-reward RL \\ 
        LEACL w/o ACL & $\circ$ & $\times$ & Sparse-reward RL \\
        LEAGUE & $\circ$ & $\times$ & Dense-reward RL \\
        Human curriculum & $\circ$ & $\circ$ & RL + Human + ACL \\
        LEACL & $\circ$ & $\circ$ & RL + LLM + ACL \\
        \bottomrule
    \end{tabular}
    \caption{An overview of the components used by LEACL and baseline algorithms for task decomposition, parameterized predicates, and specifications in RL.}
    \label{tab:components}
    \vspace{-10pt}
\end{table}
\section{Results}
Table \ref{tab:results} presents the evaluation results of all algorithms. LEACL outperformed the sparse reward and LEACL w/o ACL baselines across all tasks by a large margin, and outperformed LEAGUE in most tasks, demonstrating the effectiveness of utilizing both LLMs with ACL. LEACL also showed competitive performance and even compared to the best-performing human curriculum. 

Our key observations are summarized below. 
\paragraph{Task decomposition alone is insufficient for RL to solve even short-horizon tasks}
While task decomposition introduces intermediate feedback that can aid in solving manipulation tasks, we find that it alone is insufficient for reinforcement learning to master certain subtasks, even when their horizons are relatively short. For instance, consider the decomposition of a task where the agent first needs to \textit{reach the drawer} and then \textit{grasp the handle}. Although the former subtask positions the gripper near the drawer, the latter remains difficult to accomplish with only the sparse reward of task completion. This observation is reflected in our empirical results, where the \textit{LEACL w/o ACL} baseline fails on nearly all evaluation tasks, achieving a modest success rate of only 32.8\% on the relatively simpler task \textit{Put the white bowl on the plate}.

\paragraph{Designing shaping rewards for long-horizon manipulation tasks is challenging}
Designing effective shaping rewards for manipulation tasks poses several practical challenges, even for human experts. 

First, constructing dense rewards often requires extensive task-specific tuning, and their effects can vary significantly across objects due to differences in geometry, dynamics, and physical properties. For example, a seemingly minor change—such as defining the lift reward based on the gripper’s height versus the object’s height—can result in noticeably different behaviors. Furthermore, a reward function that is effective for one object (\eg, a bowl) may be suboptimal or even counterproductive for another (\eg, a ketchup bottle), limiting the generality of dense rewards across diverse tasks.

Second, appropriately balancing dense shaping rewards with task completion reward is non-trivial. Overweighting dense rewards can lead the agent to prioritize intermediate progress while neglecting task success, ultimately reducing performance. This trade-off is reflected in our experiments, where LEAGUE, which incorporates dense shaping rewards, significantly underperforms LEACL, which relies solely on sparse completion rewards, on 4 out of 5 tasks. In terms of policy quality, we also observe that sparse task completion rewards—particularly when combined with early termination upon success—yield more precise and reliable manipulation behaviors. In contrast, dense rewards often encourage ``sloppy" behaviors that make incremental progress but lack the precision needed to satisfy all task predicates, especially in tasks requiring conjunctions of predicates. 

Overall, while shaping rewards can accelerate early-stage learning, their design complexity, variability across tasks, and unintended side effects make them less suitable for learning generalizable policies in long-horizon, multi-step manipulation scenarios.

\paragraph{LLMs rival or even surpass human experts in curriculum generation}
We also compare the capabilities of LLMs in task decomposition and meta-task generation with those of human experts. The human curriculum baseline replaces LLMs with humans in both task decomposition and meta-task generation. Human curriculum still outperforms LEACL in 3 out of 5 tasks. However, to achieve this performance gain, a human expert requires several iterations to design the subtasks and meta-task spaces. In contrast, we observe that a well-designed prompt enables LLMs to decompose tasks and create effective task spaces in a zero-shot manner without any supervision effort across different manipulation tasks.








\section{Conclusions and Future Work}
In this paper, we introduce LEACL, designed to address long-horizon manipulation problems with the assistance of LLMs. Unlike prior approaches that predominantly leverage LLMs for direct curriculum generation or dense reward functions, LEACL uses LLMs to produce crucial task-dependent specifications, including parameterized task spaces and difficulty measures. Our empirical evaluations across five challenging manipulation tasks demonstrate LEACL’s efficacy, showcasing superior asymptotic success rates after convergence and a significant reduction in human supervision.

A major limitation of LEACL is its reliance on a predefined grammar—a set of predicates used to decompose a given task—provided a priori, such as the complete predicate set available in LIBERO+. Future work could explore methods to relax this dependency by leveraging LLMs to automatically propose task-specific grammars.


\bibliographystyle{IEEEtran}
\bibliography{references}

\end{document}